\documentclass{ieeeaccess}
\usepackage{arydshln}
\usepackage{adjustbox}
\usepackage{cite}
\usepackage{colortbl}
\usepackage{textcomp}
\usepackage{placeins}
\usepackage{multirow}
\usepackage{amsmath,amssymb,amsfonts}
\usepackage{algorithmic}
\usepackage{algorithm}
\usepackage{subcaption}
\usepackage{textcomp}
\usepackage{placeins}
\usepackage{lipsum}
\usepackage{mathtools}
\usepackage{cuted}
\usepackage[dvipsnames]{xcolor}
\usepackage{times}
\usepackage{epsfig}
\usepackage{array,multirow,graphicx}
\usepackage{epstopdf}
\usepackage{enumitem}
\usepackage{fixltx2e}
\usepackage{array}
\usepackage{comment}
\usepackage{pifont}

\usepackage[pagebackref=true,breaklinks=true,letterpaper=true,colorlinks,bookmarks=false]{hyperref}

\def\BibTeX{{\rm B\kern-.05em{\sc i\kern-.025em b}\kern-.08em
    T\kern-.1667em\lower.7ex\hbox{E}\kern-.125emX}}
    \newcommand{\xmark}{\ding{53}}%
\begin{document}

\doi{}

\title{Robust Image Classification Using A Low-Pass Activation Function and DCT Augmentation}
\author{\uppercase{Md Tahmid Hossain}\authorrefmark{1}, 
{SHYH WEI TENG\authorrefmark{1}, 
FERDOUS SOHEL\authorrefmark{2}, \IEEEmembership{Senior Member, IEEE}, AND 
GUOJUN LU\authorrefmark{1}}, 
\IEEEmembership{Senior Member, IEEE}
}
\address[1]{School of Engineering, Information Technology and Physical Sciences, Federation University, VIC 3842, Australia }

\address[2]{Discipline of Information Technology, Murdoch University, Perth, WA 6150, Australia}

\tfootnote{Federation University Australia Research Grants fund this work.}

\markboth
{Hossain \headeretal: Robust Image Classification Using A Low-Pass Activation Function and DCT Augmentation}
{Hossain \headeretal: Robust Image Classification Using A Low-Pass Activation Function and DCT Augmentation}

\corresp{Corresponding author: Md Tahmid Hossain (email: mt.hossain@federation.edu.au).}

\begin{abstract}
Convolutional Neural Network's (CNN's) performance disparity on clean and corrupted datasets has recently come under scrutiny. In this work, we analyse common corruptions in the frequency domain, i.e., High Frequency corruptions (HFc, e.g., noise) and Low Frequency corruptions (LFc, e.g., blur). Although a simple solution to HFc is low-pass filtering, ReLU -- a widely used Activation Function (AF), does not have any filtering mechanism. In this work, we instill low-pass filtering into the AF (LP-ReLU) to improve robustness against HFc. To deal with LFc, we complement LP-ReLU with Discrete Cosine Transform based augmentation. LP-ReLU, coupled with DCT augmentation, enables a deep network to tackle the entire spectrum of corruption. 
We use CIFAR-10-C and Tiny ImageNet-C for evaluation and demonstrate improvements of 5\% and 7.3\% in accuracy respectively, compared to the State-Of-The-Art (SOTA). We further evaluate our method's stability on a variety of perturbations in CIFAR-10-P and Tiny ImageNet-P, achieving new SOTA in these experiments as well. {To further strengthen our understanding regarding CNN's lack of robustness, a decision space visualisation process is proposed and presented in this work.}

\end{abstract}


\titlepgskip=-15pt

\maketitle

\begin{figure*}
\begin{center}
   \includegraphics[width = .9\linewidth]{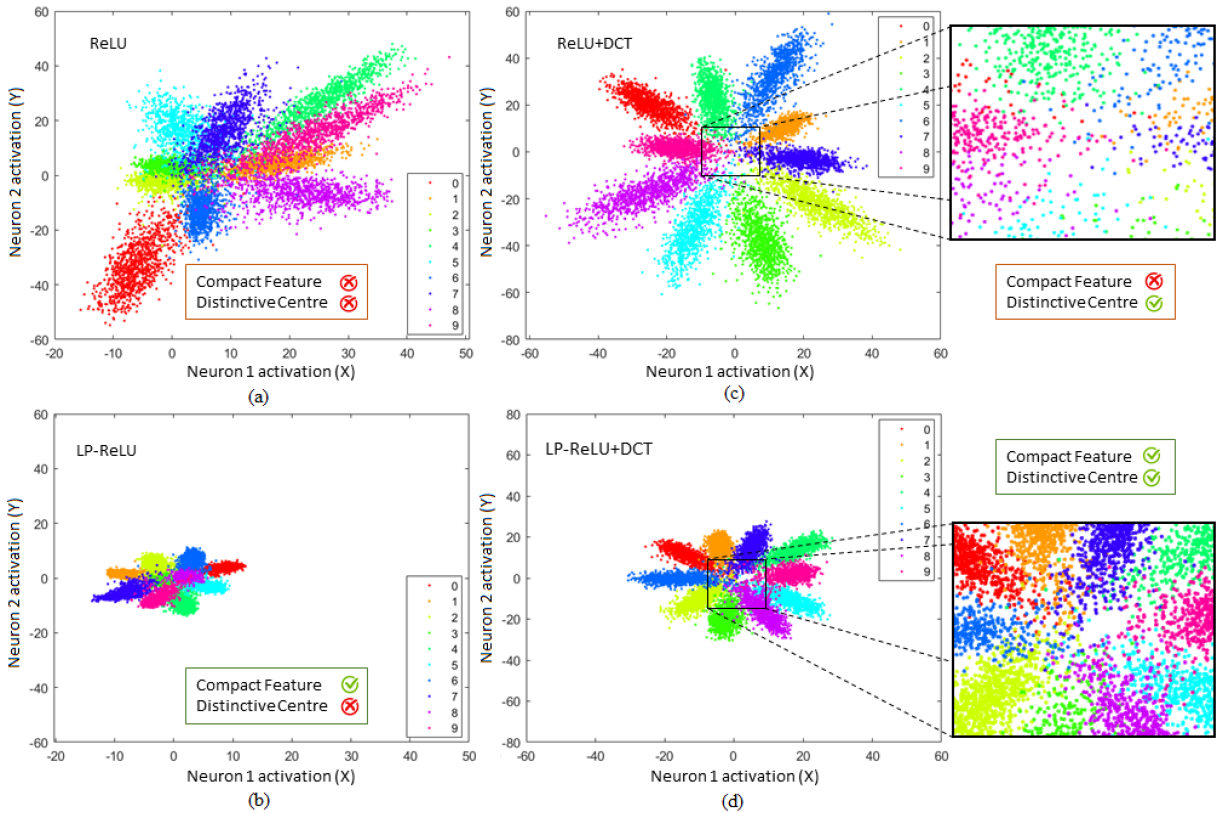}
\end{center}
   \caption{{Visualising deep feature characteristics for different network setting on the MNIST dataset. }\textbf{(a)} ReLU allows activations or features to flow through network layers without any filtering. This introduces sparsity in the feature space. Such sparsity can cause shift in the feature space in the presence of data corruption, even when corruption is visually imperceptible. \textbf{(b)} Our proposed AF has low-pass filter built-in, which enforces feature compactness. This, in effect, limits the internal feature shift and resists corrupted features from drifting away. \textbf{(c)-(d)} As discussed later in Section \ref{subsec:frequency}, robustness against LFc demand better distinction for weak features (weak features reside in the centre of the above plots). DCT data augmentation \cite{hossain1} improves robustness for both AFs, i.e., ReLU \textbf{(c)}, and LP-ReLU \textbf{(d)}- especially against LFc by increasing the inter-class distance (distance among weak features) in the centre. See the zoomed insets for comparison. Despite overall improvement, feature sparsity still exists in ReLU \textbf{(c)}, which makes it vulnerable to corruptions. { All plots in this figure use features taken from a CNN proposed to visualise deep features and decision boundary (details in Figure \mbox{\ref{fig:visualize}}, Section \mbox{\ref{sec:visualize}})}.
}
\label{fig:main}
\end{figure*}

\section{Introduction}
\IEEEPARstart{D}{eep} Convolutional Neural Networks (CNNs) {achieve} high classification and recognition accuracy on i.i.d. (independent and identically distributed) {benchmark datasets}. However, performance deteriorates on corrupted datasets even when the corruption is visually imperceptible. This is particularly concerning for real-world safety-critical applications where the i.i.d. assumption does not always hold. For example, a self-driving car should not presume to have ideal driving conditions at all times. Optical sensors are expected to encounter a wide range of corruptions stemming from trivial physical phenomena (e.g., signal noise, motion, and defocus blur), variable weather conditions (e.g., brightness, snow, frost, and fog), and digital artefacts (e.g., pixelate effect, and data compression). Therefore, accounting for corruptions at inference time and closing the performance gap between clean and corrupted datasets are important.

Different data augmentation techniques \cite{autoaugment1,hossain1,patchG1,augmix,deepAugment} are widely used to address this issue. In addition to data augmentation, adversarial training \cite{HendrycksALP1,carlini2017adversarial,ford2019adversarial,athalye2018obfuscated,Xie}, self-supervised learning \cite{HendrykRotation01,selfsuper2}, domain adaptation \cite{Benz,Schneider}, and loss function optimization \cite{madry2017towards,HendrykRotation01} are some of the other ways explored to enhance CNN's robustness against such corruptions.

In this work, we take a different approach and investigate corruptions from the frequency domain. We emphasise on the role of Activation Functions (AFs) and argue that if designed properly, AFs can substantially enhance CNN's robustness against corruptions. 
Currently, ReLU \cite{alexnet1} is the most widely used AF because of its computational efficiency and convergence ability compared to other AFs \cite{malaysia1}. ReLU does not suffer from the vanishing gradient problem \cite{vanishing1} and it allows very deep networks to be optimised through backpropagation. However, ReLU and a number of its variants lack robustness against common corruptions (as shown later in Section \ref{sec:results}). 

By definition, ReLU blocks out any negative input but does not alter positive input at all. Therefore, the intermediate layer features inside ReLU-based networks become sparse. Sparse features maximise the inter-class distance to gain high classification accuracy but could compromise the classification performance with their higher intra-class dispersion (see Figure \ref{fig:main}(a)). This compromise does not harm CNN's overall performance so long as the evaluation dataset is i.i.d. or clean. However, the compromise (i.e., feature sparsity) gets exposed in the presence of input corruptions - especially High Frequency corruptions or HFc (e.g., Gaussian or speckle noise). HFc, even when visually imperceptible, can cause the corrupted data to drift away to a different part of the feature space \cite{goodfellow1} in the absence of a low-pass filter. This leads to misclassification.

The classic signal processing fix to a HFc is low-pass filtering. For example, voice recorders use low-pass filtering to cancel the `hiss’ noise originating from electromagnetic interference. As shown by Campbell et al., \cite{campbell1968}, Human Visual System (HVS) also has limited sensitivity to abrupt or high frequency changes owing to an inherent low-pass filtering process. This particular property of HVS is later exploited in JPEG image compression as well \cite{wallace1}.
Inspired from these observations, we incorporate a low-pass filtering mechanism inside the proposed AF, namely, Low-Pass ReLU (LP-ReLU) and show its efficacy against corruptions -- especially HFc. LP-ReLU resists features from drifting, even when there is HFc by enforcing a compact feature space (see Figure \ref{fig:main}(b)).

As for Low Frequency corruptions (LFc), one might think of AFs with high pass filtering as a potential fix. However, corrupted features from LFc substantially overlap with meaningful image features (details in Section \ref{sec:proposed}). Using a high pass filter can exacerbate the problem by discarding meaningful features along with the corrupted ones. Therefore, LFc demand another way around.  
We find that a data augmentation method based on Discrete Cosine Transform (DCT) \cite{hossain1} can further boost overall robustness, especially against LFc. DCT augmentation provides greater distinction for weak features (at the centre region of the feature space) with LFc (see Figures \ref{fig:main}(c) and \ref{fig:main}(d)). Feature characteristics for both HFc and LFc are discussed in details in Section \ref{sec:proposed}.

We evaluate our method's efficacy on benchmark datasets (CIFAR-10-C and Tiny ImageNet-C \cite{HendrycksALP1}) containing common corruptions. We further conduct performance stability tests on perturbed datasets (CIFAR-10-P and Tiny ImageNet-P \cite{HendrycksALP1}). Experimental results show that LP-ReLU provides better robustness and stability against corruptions compared to contemporary works.

{We also stress the importance of visualisation of the learned decision space to better understand the dynamics of CNN's robustness against common corruptions. To this end, we propose a way of visualising the learned decision space and how a corrupted input triggers misclassification. We show that the learned decision boundaries extend to infinity and extrapolate predictions in the space beyond the scope of the data manifold.} This observation reiterates the importance of a compact feature space to suppress the detrimental effects of corruptions.

To summarise, {we make the following contributions:}

\begin{itemize}
    \item {We analyse image corruptions from a frequency perspective and show that CNN's weakness against corrupted data can be addressed by using appropriate AFs and augmentation techniques.}
    \item We propose novel AFs (LP-ReLU) by embedding low-pass filtering properties into ReLU. 
    \item To Complement LP-ReLU in tackling LFc, we use DCT based augmentation \cite{hossain1}.
    \item {For an understanding of CNN's overall robustness against data corruptions, we propose a method to visually illustrate CNN's decision boundaries and its intermediate feature space.}  
    \item Finally, we extensively evaluate the accuracy and stability of our proposed method on benchmark datasets and demonstrate the efficacy of our method in improving CNN's robustness.
\end{itemize}

The rest of the paper is organised in the following way: Section \ref{sec:related_work} discusses the related work. The proposed AFs and the recommended augmentation methodology are discussed in Section \ref{sec:proposed}. Section \ref{sec:results} provides details on the benchmark datasets, extensively evaluates the performance of different networks, and outlines the implementation details. Section \ref{sec:visualize} presents the proposed decision space visualization process. Finally, Section \ref{sec:conclusion} concludes the paper.

\section{Related Work}\label{sec:related_work}
\subsection{Robustness Against Distortion}
Although a sizable body of work is available on adversarial examples \cite{goodfellow1,carlini2017adversarial}, i.e., images which are purposely generated to fail the network, common form of corruptions (e.g., image noise, blur, fog and frost) have received limited attention. {Real-world applications are susceptible to such deformations and as argued in \mbox{\cite{HendrycksALP1}}, they deserve separate attention.}

Dodge et al. \cite{icccn1} study the impact of image quality degradation and report that CNN's performance starts deteriorating even when the corruption is visually imperceptible. They train an ensemble of networks \cite{akaram1} where each network becomes expert on a specific degradation. Hossain et al. \cite{hossain1} report similar findings and propose a DCT based data augmentation technique to ameliorate robustness. 

Upon realising CNN's susceptibility to input corruptions, Hendrycks et al. \cite{HendrycksALP1} has recently published a range of benchmark datasets with 19 common corruptions. A number of works \cite{HendrycksALP1,ford2019adversarial} explored adversarial training to strengthen CNN's robustness, but Rusak et al. \cite{rusak2020increasing} argue that performance gain achieved through adversarial training do not translate consistently to common corruptions. Rather, they propose a data augmentation technique where generative networks are used to produce a wide array of noisy images. Performance improvement is observed on most of the noisy and some of the blur type corruptions. 

AutoAugment \cite{autoaugment1} is another augmentation technique that was initially designed to boost clean performance. It incorporates a method to dynamically choose from a pool of image processing policies during training. Recently, AutoAugment is found to be effective in improving CNN's robustness against common corruptions as well \cite{fourier1,augmix}. Hendrycks et al. \cite{augmix} build on the work of AutoAugment and propose an improved augmentation technique called AugMix. In AugMix, a chain of augmentation techniques is applied to an image and layered together. These images are argued to be more natural and close to the original images, unlike some of the other works \cite{patchG1,cutmix1,mixup1,cutout1}. It is worth noting that some of the augmentation techniques overlap with the test set. For instance, Gaussian and Speckle noise augmentations appear in \cite{rusak2020increasing,zhou1} despite their presence in the test data. Similarly, posterisation processing used in \cite{augmix} visually resembles one of the corruptions (JPEG compression) present in the test set. Such overlaps lead to performance gain in the corresponding corruption category. However, this gain does not always generalise well to unseen corruptions as outlined in \cite{geirhos1}.

Zhang et al. \cite{zhang2019making} argue that the conventional Max-Pooling violates Nyquist Sampling theorem resulting in enhanced sensitivity to corruption. As a fix, blurred pooling is used right before down-sampling. 

Self-supervised training, in addition to conventional supervised training, has been reported to yield promising results in \cite{HendrykRotation01}. An auxiliary 4-way head is trained to predict the rotation angle of an input which is found to improve overall robustness as well. Sun et al. \cite{selfsuper2} further improve on \cite{HendrykRotation01} and enable test time parameter update, i.e., online learning.

Some of the recent works \cite{Benz,Schneider} treat CNN's lack of robustness as a domain adaptation problem and propose to use rectified Batch Normalization Statistics (BNS) \cite{Benz}.


\begin{figure*}
\begin{center}
   \includegraphics[width=.95\linewidth]{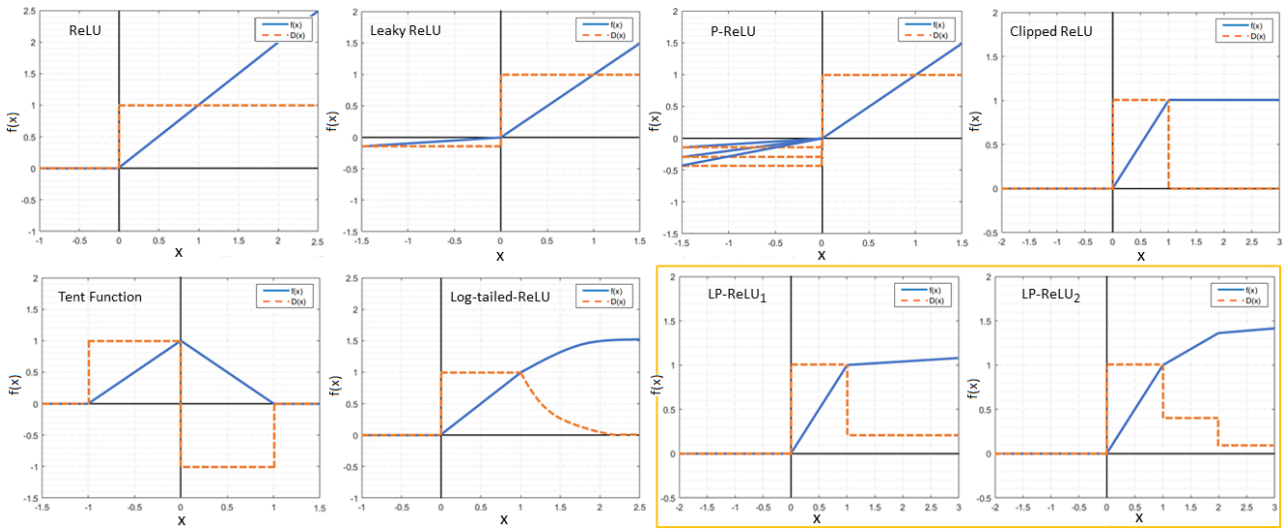}
\end{center}
   \caption{Activation Functions $f(x)$ and their corresponding $1^{st}$ derivatives $D(x)$. The proposed variants of Activation Functions, i.e., LP-ReLU\textsubscript{1} and LP-ReLU\textsubscript{2} are highlighted in the lower right panel.}
\label{fig:AFs}
\end{figure*}

\subsection{Activation Functions}\label{sec:AFs}
 
 Adding non-linearities in the form of AFs has been an integral part of deep CNNs learning process. In the early phase, Sigmoidal functions were widely used but later got sidelined as they suffer from the vanishing gradient problem \cite{vanishing1}. Sigmoidal functions squash values within a finite range (typically $[0,1]$ or $[-1,1]$) which leaves saturation points on both sides making it impossible to find a slope in these regions. This halts the error backpropagation, and the network struggles to reach the global minima. This issue becomes particularly severe in very deep networks where the loss has to travel back a long way to have meaningful learning. Rectified Linear Unit or ReLU does not suffer from the vanishing gradient problem and has long been the `de-facto' AF in deep networks. We will next discuss the pros and cons of different AFs found in the literature.\\
 \textbf{ReLU. }
 ReLU \cite{relu1Hinton,relu1} is a piece-wise linear monotonic function. It simply does not have any response on the negative side, and any positive value remains unchanged (Equation \ref{equRelu}).
 ReLU is easy to compute \cite{alexnet1}, a derivative is available everywhere along with the positive range, which makes it a great choice for deep networks. It is also hypothesised to mimic the biological neuron firing process to justify its effectiveness. This correlation, however, is disputed in \cite{li2016sparseness} based on the IFM (Integrate and Fire Model \cite{relu-father}) in biological neuron. Unlike ReLU where the output slope is always constant ($45^{\circ}$, see Figure \ref{fig:AFs}), according to IFM, the biological neuron's output slope varies depending on the resistance present in the cell membrane. We take this as an inspiration for our proposed AF's design in Section \ref{sec:proposed}.
  \begin{equation}\label{equRelu}
 f(x)=\max (0, x)
 \end{equation} 
As shown later in Section \ref{sec:shift}, ReLU's unbounded nature leads to feature sparsity and lets small corruptions to cause the feature of its data to shift significantly away from their clean version, even when corruption is negligible. This results in misclassification. In simple words, ReLU allows CNNs to achieve impressive accuracy on clean set but makes CNN vulnerable to data corruptions. \\
 \textbf{Leaky ReLU and Parametric ReLU. }
 Because ReLU has a zero response for negative values, a large number of neurons may never fire in the absence of proper initialisation, and hyper-parameter setup \cite{leaky1} resulting in `dead neurons'. Leaky ReLU has been proposed \cite{leaky1} to address the dead neuron problem by introducing a small constant slope ($\alpha = .01$) on the negative side (Equation \ref{equLeaky}). Leaky ReLU has a derivative on both sides of the origin (Figure \ref{fig:AFs}). In Parametric ReLU (P-ReLU) \cite{prelu1}, the slope $\alpha$ is set up as a learnable parameter rather than a constant and accuracy improvement is reported \cite{prelu1}. However, leaky variants do not consistently outperform vanilla ReLU across datasets.
 \begin{equation}\label{equLeaky}
 f(x)=\left\{\begin{array}{ll}
x, & x>0, \\
\alpha x, & \text { otherwise }
\end{array}\right.
\end{equation}
\textbf{Clipped ReLU. }
Clipped ReLU (C-ReLU)\cite{clipped1} is simply vanilla ReLU clipped to a constant based on a threshold $A$ (Equation \ref{equBounded}). Clipped ReLU was initially proposed in speech recognition \cite{clipped1} to increase training stability by avoiding gradient explosion. Later, it found its way into computer vision and other domains of deep learning as well. For example, Clipped ReLU's finite output range is found to be handy in applications where number representation capability is limited \cite{malaysia1}. Despite Clipped ReLU's different set of use-cases, it incorporates a hard low pass filtering which leaves large saturation points in the bounded region (see Figure \ref{fig:AFs}). Consequently, vanishing gradient remains a problem, and there is no way to discriminate features beyond the threshold $A$.
 \begin{equation}\label{equBounded}
f(x)=\left\{\begin{array}{ll}
\max (0, x), & 0 < x \leq A,\\
A, & x>A
\end{array}\right.
\end{equation}
\textbf{Tent Function. }
Tent AF \cite{tent1} is built from two ReLU units and it is symmetric around the origin (Equation \ref{equTent}). It is designed to resist adversarial attacks but unlike ReLU, this AF is not monotonic and has large saturation regions on both sides beyond a threshold $\delta$ (see Figure \ref{fig:AFs}).
\begin{equation}\label{equTent}
f(x ; \delta)=\max (0, \delta-|x|)
\end{equation}
\textbf{Log-tailed ReLU. }
As the name suggests, Log-tailed ReLU \cite{logtail1} is identical to ReLU up to a threshold $A$. The growth of the function is logarithmic thereafter $[A,\infty)$. However, this tail part rapidly converges to clipped ReLU with increasing $x$ as can be seen from Equation \ref{equlog} and Figure \ref{fig:AFs}.
\begin{equation}\label{equlog}
f(x)=\left\{\begin{array}{cl}
0, & x \leq 0,\\
x, & 0 < x \leq A, \\
A + \log (x-A), & x > A \\
\end{array}\right.
\end{equation}
\textbf{Tanh Function. }
Hyperbolic tangent function or $tanh$ is sigmoidal in shape and squashes values inside a finite range of $[-1,1]$ (Equation \ref{equtanh}). $tanh$ is centred around the origin. It has faster convergence compared to other sigmoidal functions because of the wider output range \cite{lecun1998gradient}. However, it still has large saturated regions on both sides of the origin.

\begin{equation}\label{equtanh}
f(x)=\tanh (x)=\frac{\sinh (x)}{\cosh (x)}=\frac{1-e^{-2 x}}{1+e^{-2 x}}
\end{equation}
\textbf{Swish. } This activation function \cite{swish} is represented by Equation \ref{swish}. Here $\beta$ is either a constant or a trainable parameter which defines the final shape of the function. For example, when $\beta$ is close to $0$, Swish becomes a scaled linear function $f(x) = \frac{x}{2}$. For increasingly large $\beta$, Swish starts imitating ReLU. However, regardless of the $\beta$ value, the positive range of Swish remains unbounded.

\begin{equation}\label{swish}
f(x)=x \cdot \sigma(\beta x), \quad \sigma(z)=(1+\exp (-z))^{-1}
\end{equation}

\section{Proposed Approach} \label{sec:proposed}
Achieving robustness against corruptions requires an in-depth understanding of their properties. In this section, we set the foundation for our proposed approach by analysing different corruptions from a Fourier perspective. {For all our experiments presented in Figure \mbox{\ref{fig:hist}, \ref{fig:alpha}, and \ref{fig:alphabeta}}, Wide-ResNet (WRN-40-2) is used.}
\subsection{Frequency Domain Analysis}\label{subsec:frequency}

Commonly found image corruptions can be broadly categorised into two types: HFc and LFc \cite{fourier1}. We argue that successfully handling the entire corruption spectrum, i.e., both HFc and LFc, is at the core of achieving a desired level of robustness. 

\begin{figure*}
\begin{center}
   \includegraphics[width=.9\linewidth]{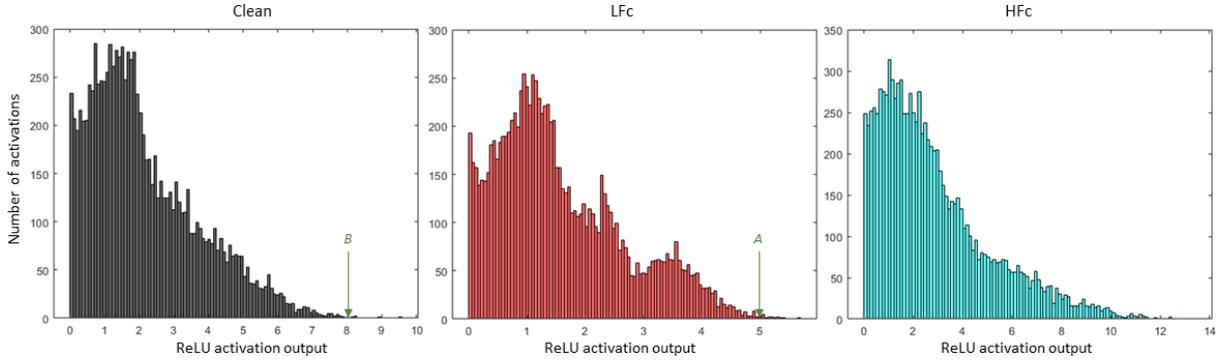}
\end{center}
   \caption{{Histograms of ReLU output, averaged across all the layers in WRN-40-2, from clean images in CIFAR-10 and Tiny ImageNet (\textbf{Left}), images with LFc (\textbf{Middle}), and images with HFc (\textbf{Right}) from CIFAR-10-C and Tiny ImageNet-C.} Horizontal axis denotes the magnitude of a feature or activation and the vertical axis denotes the number of activations having a certain magnitude. These plots show that the HFc indeed induce hyperactive features with large output magnitude (\textbf{Right}) and LFc induce relatively weak features (\textbf{Middle}). The histogram with weak features (LFc) overlaps with the clean one (\textbf{Left}) and hence LFc are harder to distinguish. These plots on clean, LFc and HFc also provide a heuristic on the potential cut-off point initialisation values ($A$ and $B$) in the proposed LP-ReLU as discussed in Section \ref{sec:cutoff}. }
\label{fig:hist}
\end{figure*}

\begin{figure}
\begin{center}
   \includegraphics[width=.9\linewidth]{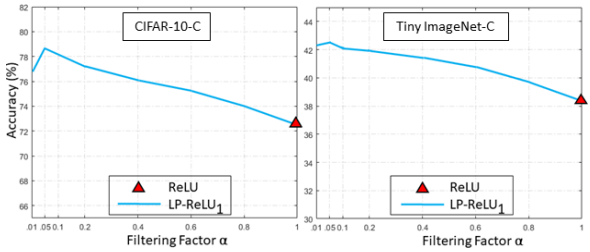}
\end{center}
   \caption{Empirical analysis for a suitable $Filtering$ $Factor$ $\alpha$ in LP-ReLU\textsubscript{1}. Notice that $\alpha \approx .05$ yields the best results and $\alpha = 1$ simply replicates the results of ReLU. {WRN-40-2 is used as the backbone for these experiments.}}
\label{fig:alpha}
\end{figure}

Noise introduces sudden spikes in the signal and thereby falls in the HFc category \cite{fourier1}. Blur, on the other hand, is another common form of corruption that removes high frequency information and thereby falls in the LFc category (e.g. Gaussian blur) \cite{fourier1}.\\
\textbf{High Frequency Corruptions. }
In signal processing applications, low-pass filtering is a common operator to remove nuisance frequencies beyond a cut-off threshold. From experimental analysis, we observe that an image with HFc induces activations with a higher magnitude compared to its clean counterpart (see Figure \ref{fig:hist}). In other words, convolution kernels fire strongly on `noisy' parts of an input with HFc. These hyperactive features coming out of the noise rather than the signal are also fed to ReLU. As ReLU does not filter input in the positive domain, i.e., for $x>0$, such features are allowed to flow through the network layers. Consequently, these corrupted features exploit the sparsity offered by ReLU and trigger heavy shift in the feature space. This way features easily end up on the wrong side of the decision boundary, causing misclassification (details in Section \ref{sec:shift}). \\
\textbf{Low Frequency Corruptions. }
As stated earlier, LFc do not have an easy filtering fix. The underlying reasons can be better understood if we move from spatial to frequency domain. 

As observed in Figure \ref{fig:hist}, convolution kernels fire weakly on images with LFc as such corruptions usually blunt the visual features. Contrary to HFc, this implies that the LFc induce relatively weaker features\footnote{By weak activation or feature, we refer to the smaller magnitude of absolute values produced from each convolution operation.} that congregate in the centre region of the feature space (as shown in Figure \ref{fig:main}). For example, a vertical edge detector kernel would produce strong output features each time it convolves over a sharp vertical edge present in an image. However, the same kernel would produce weaker features if the vertical edge is not sharp enough (e.g., loss of sharpness due to LFc).

It is known from DCT analysis that most of the energy (of a natural image) is concentrated on the low frequency spectrum \cite{wallace1}. Unlike HFc, features from LFc substantially overlap with meaningful features. Consequently, the classic fix, i.e., using a high-pass filter, is not a viable option as such a filtering might drop corrupted features as well as legitimate ones. This loss of information would not be recoverable and result in performance deterioration.

We argue that a data augmentation technique capable of mimicking LFc can significantly boost robustness not only against LFc but also against HFc.


\subsection{ Proposed Activation Function: LP-ReLU}
 
 Inspired by the HVS and traditional signal processing fix for HFc, we design two variants of low pass filters for our proposed AFs. Before explaining further, there are a couple of key differences between the proposed and traditional low-pass filtering we would like to highlight: 
 
 \begin{itemize}
     \item Unlike the traditional fix where everything beyond a cut off frequency is completely ignored, we use a soft filtering technique with a signal attenuation factor that we call $Filtering$ $Factor$. Note that completely cutting off the signal beyond a threshold would resurface the vanishing gradient problem (similar to Clipped-ReLU).
     \item In a traditional low-pass filtering operator, a cut-off frequency is chosen based on the maximum frequency available in the signal of interest (following Nyquist Theorem  \cite{nyquist1}). However, computing such a cut-off frequency is complicated for visual datasets like images. Hence, we design two low pass filter variants in our AF namely LP-ReLU\textsubscript{1} (one cut-off point) and LP-ReLU\textsubscript{2} (two cut-off points) with different cut-off point selection strategies.
 \end{itemize}

\begin{figure}
\begin{center}
   \includegraphics[width=.7\linewidth]{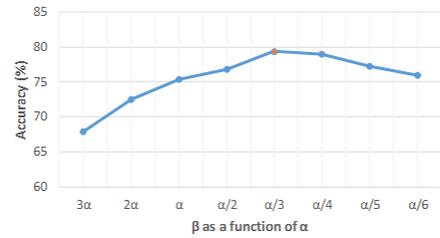}
\end{center}
   \caption{{Empirical analysis of $\beta$ as a function of decreasing $\alpha$ in \mbox{LP-ReLU\textsubscript{2}}. Here, $\alpha$ is the phase 1 and $\beta$ is the phase 2 $Filtering$ $Factor$. In this experiment, WRN-40-2 is used as the backbone and CIFAR-10-C test fold as the dataset. This experiment serves as a heuristic to initialise $\beta$ regardless of the dataset to avoid data dependency.}}
\label{fig:alphabeta}
\end{figure}

\subsection{LP-ReLU\textsubscript{1}} 
Equation \ref{eq:1} represents the first variant of our proposed AF, i.e., LP-ReLU\textsubscript{1}. A closer look into Equation \ref{eq:1} reveals that, LP-ReLU\textsubscript{1}, upto a threshold $A$, is equivalent to ReLU. Beyond this threshold, the input features get attenuated by a $Filtering$ $Factor$ $\alpha$ ($\alpha \in [0,1]$). Note that $\alpha = 0$ will make LP-ReLU\textsubscript{1} equivalent to Clipped ReLU. $\alpha = 1$, on the other hand, would make LP-ReLU\textsubscript{1} equivalent to ReLU. 
    
\begin{equation}\label{eq:1}
F(x)=\left\{\begin{array}{ll}
0, & x \leq 0, \\
x, & x \in (0,A], \\
A + \alpha (x-A), & x > A  \\
\end{array}\right.
\end{equation}

Equation \ref{eq:2} denotes the derivative of LP-ReLU\textsubscript{1} at each piece-wise linear stage. It is evident that LP-ReLU\textsubscript{1} has a slope at every point beyond the origin, which is a much-desired property for training any neural network. See the graphical representations of LP-ReLU\textsubscript{1} and its derivative in Figure \ref{fig:AFs}.

\begin{equation}\label{eq:2}
D(x)=\left\{\begin{array}{ll}
0, & x \leq 0, \\
1, & x \in (0,A], \\
\alpha,  & x > A \\
\end{array}\right.
\end{equation}

\subsection{LP-ReLU\textsubscript{2}} 
In contrast to LP-ReLU\textsubscript{1}, LP-ReLU\textsubscript{2} has two cut-off points $A$, and $B$ ($A < B$) in series and two corresponding $Filtering$ $Factors$ $\alpha$, and $\beta$ ($\alpha > \beta$) as can be seen from Equation \ref{eq:3}. The graphical representations of LP-ReLU\textsubscript{2} and its derivative are shown in Figure \ref{fig:AFs}.

LP-ReLU\textsubscript{2}'s effectiveness against corruptions can be attributed to the following reasons:

\begin{itemize}
    \item \textbf{Phase 1 (soft) filtering. } A relatively larger $\alpha$, i.e., $\alpha \rightarrow 1$ (for $x \in (A,B]$) allows greater feature sparsity in the center, i.e., sparsity for weak features in the center region where LFc congregate (as shown in Figure \ref{fig:main}). This initial soft filtering gives our network enough sparsity in the centre to accommodate weak features in distinct regions. 
    \item \textbf{Phase 2 (hard) filtering. } A relatively smaller $\beta$, i.e., $\beta \rightarrow 0$ (for $x > B$) means hard suppression for large $x$, i.e., towards the perimeter of the feature space. This hard filtering stage suppresses noise in the signal, and limits feature shift by ensuring a compact feature space. 
\end{itemize}

\begin{equation}\label{eq:3}
F(x)=\left\{\begin{array}{ll}
f\textsubscript{1}(x) = 0, & x \leq 0, \\
f\textsubscript{2}(x) = x, & x \in (0,A], \\
f\textsubscript{3}(x) = A + \alpha (x-A), & x \in (A,B], \\
f\textsubscript{4}(x) = max(f\textsubscript{3}(x)) + \\
\quad \quad \quad \quad \beta (x-max(f\textsubscript{3}(x)), & x > B   \\
\end{array}\right.
\end{equation}


Equation \ref{eq:4} represents the derivative of LP-ReLU\textsubscript{2}. Much like LP-ReLU\textsubscript{1}, LP-ReLU\textsubscript{2} also has a slope at every piece-wise linear stage.

\begin{equation}\label{eq:4}
D(x)=\left\{\begin{array}{ll}
0, & x \leq 0, \\
1, & x \in (0,A], \\
\alpha,  & x \in (A,B], \\
\beta, & x > B   \\
\end{array}\right.
\end{equation}

Overall, LP-ReLU\textsubscript{2} shows us that successfully handling the entire corruption spectrum requires separate approaches, i.e., sparsity for LFc and compactness for HFc. As shown later in this work, LP-ReLU\textsubscript{2} has a slight advantage over LP-ReLU\textsubscript{1} in accuracy but LP-ReLU\textsubscript{1} is faster to train.

\subsection{Cut-off Point and Filtering Factor} \label{sec:cutoff}

For LP-ReLU\textsubscript{1}, the cut-off value $A$ is set as a learnable parameter initialized with the value 6 based on heuristic \cite{alexkri1}. $Filtering$ $Factor$ $\alpha$ can be a constant or cast as a trainable parameter. In this work, based on the analysis in Figure 4, we use $\alpha = 0.05$ as it yields the best performance as a constant parameter. When cast as a trainable parameter, we encourage using $0.05$ as the initial value while making sure $\alpha < 1$ at all times.

For LP-ReLU\textsubscript{2}, we cast the cut-off values $A$ and $B$ as learnable parameters and ensure at all times during training they maintain a buffer in between, i.e., $A <B$. We investigate a histogram-based approach to initialise these hyperparameters (with $A = 5$ and $B = 8.1$). In Figure \ref{fig:hist}, we plot the ReLU output (averaged over all layers) histograms for the clean, LFc, and HFc.
It can be derived that ReLU output beyond $B$ is likely to be noise. On the other hand, output under the threshold $A$ could either be the true signal or LFc. This is why to initialise, we choose $A$ as the first cut-off point (with a soft $Filtering$ $Factor$ $\alpha$) and $B$ as the second cut-off point (with a hard $Filtering$ $Factor$ $\beta$). For the experiments in Figure \ref{fig:hist}, \ref{fig:alpha}, and \ref{fig:alphabeta}, 15\% of the train set is used as the validation subset on which the statistics are derived.

Both $Filtering$ $Factor$s, $\alpha$ and $\beta$ are learnable parameters. $\alpha$ is initialised in the same way as described in LP-ReLU\textsubscript{1}, and $\beta$ is cast as a function of $\alpha$. At all times during training, the $\alpha > \beta$ relation is maintained. From empirical analysis (Figure \ref{fig:alphabeta}) we initialise with $\beta = \frac{\alpha}{3}$. During training, we observed that the performance variance with different initialisations is marginal as long as these hyperparameters oscillate between optimal and near-optimal values. This allows the hyperparameter initialisation to be flexible and transferable to other datasets.

\begin{figure}
\begin{center}
   \includegraphics[width=.9\linewidth]{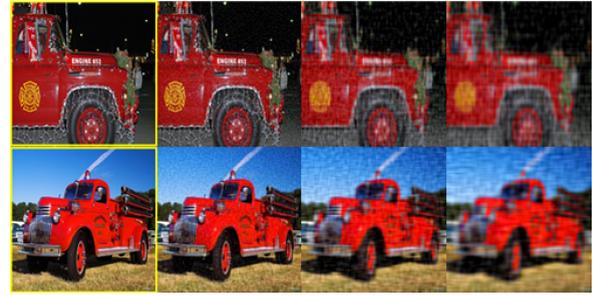}
\end{center}
   \caption{Sample images from DCT augmentation \cite{hossain1}. First image in each row is the clean one and drop of information based on DC coefficient intensifies from left to right.}
\label{fig:dct}
\end{figure}

\subsection{DCT Augmentation}

To further strengthen CNN's robustness against corruptions- especially LFc, we incorporate DCT data augmentation \cite{hossain1} along-side LP-ReLU. DCT data augmentation randomly drops low impact HF information based on DCT coefficients. Because of the randomness, some LF information gets dropped as well, which is found to be more effective as it introduces data diversity \cite{hossain1}. Unlike JPEG compression \cite{wallace1}, a block size equal to the input's spatial resolution is used. DCT data augmentation is corruption agnostic but generalises well to a wide array of common corruptions- especially LFc. The impact of DCT data augmentation coupled with LP-ReLU, can be visually observed in Figure \ref{fig:main} as the centre region (LFc) becomes much more distinct. Samples derived from DCT augmentation can be seen in Figure \ref{fig:dct}.

\begin{figure*}
\begin{center}
   \includegraphics[width=.96\linewidth]{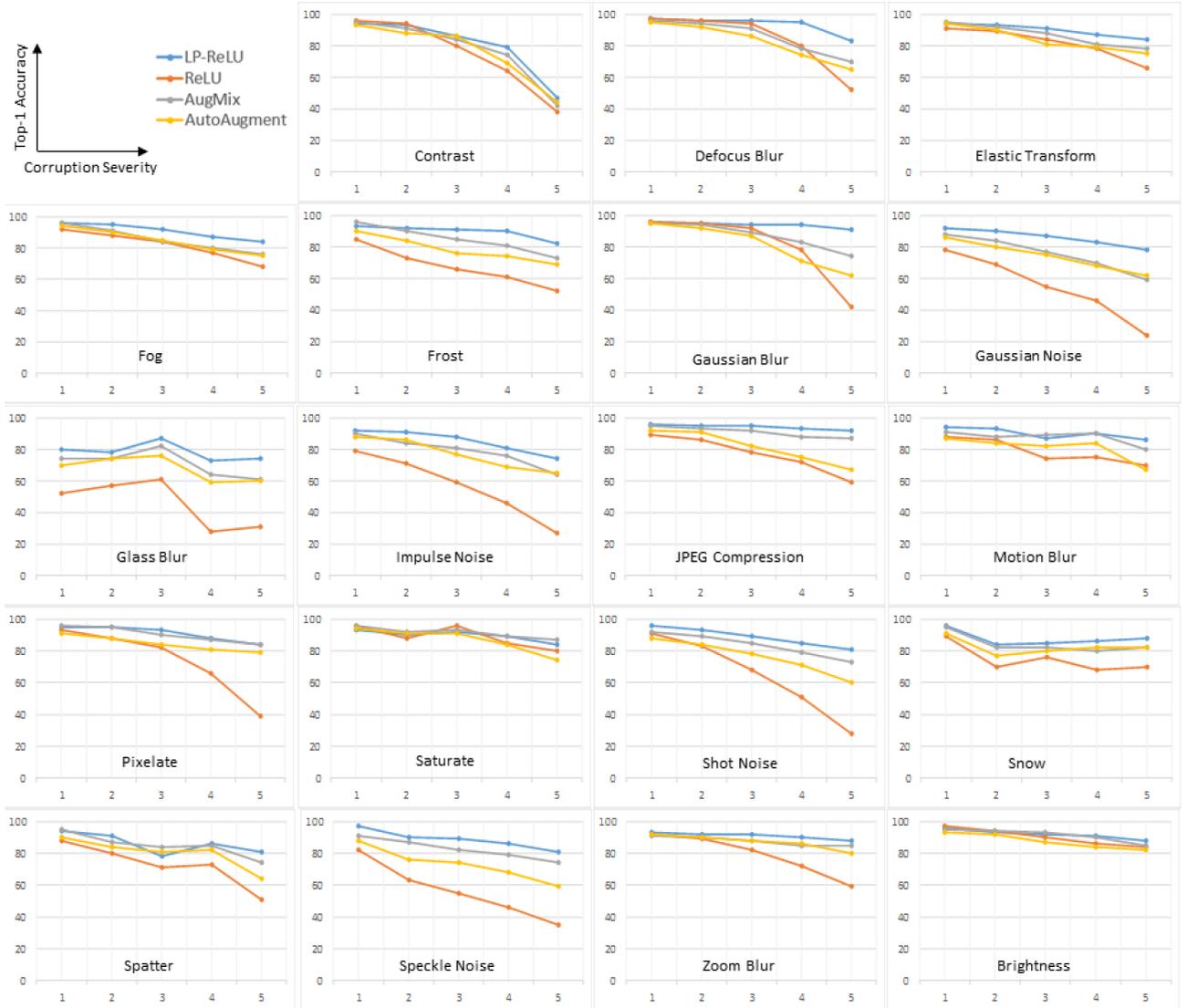}
\end{center}
   \caption{Performance comparison between LP-ReLU (LP-ReLU\textsubscript{2} + DCT) and contemporary methodologies with progressively increasing corruption in CIFAR-10-C. $1$ and $5$ denote lowest and highest level of corruption respectively. LP-ReLU exhibits consistently better performance across all severity levels.}
\label{fig:grand_collage}
\end{figure*}

\section{Performance Analysis}\label{sec:results}

 \subsection{Datasets}\label{sec:datasets}
 To evaluate robustness of a deep classifier, the following datasets are used:\\
 \textbf{CIFAR-10 and Tiny ImageNet\footnote{From here on, whenever we mention clean data set or clean accuracy, we refer to the original dataset contained in either CIFAR-10 or Tiny ImageNet.}. } CIFAR-10 \cite{cifar1} has 50,000 $32 \times 32\times 3$ clean images equally distributed in 10 classes. The split is 50,000 training images and 10,000 test images. 
 Tiny ImageNet \cite{tiny1} is an ImageNet \cite{imagenet1} subset comprising of 100,000 clean training images with 200 classes. Each image has a spatial resolution of $64 \times 64 \times 3$. Each of the classes has 500 training and 50 test images (total 10,000 test images). \\
\textbf{CIFAR-10-C and Tiny ImageNet-C. } CIFAR-10-C \cite{HendrycksALP1} is effectively the corrupted variant of CIFAR-10 test set. There are a total of 19 corruption types categorised into noise, blur, weather, and digital corruptions. Each image within a particular corruption type undergoes five increasing levels of severity to ensure a thorough evaluation. Thereby, CIFAR-10-C has $19 \times 10,000 \times 5$ or $950,000$ test images in total.

 The corruption induction process for Tiny ImageNet-C \cite{HendrycksALP1} is identical to that of CIFAR-10-C. In total, Tiny ImageNet also has $19 \times 10,000 \times 5$ or $950,000$ test images.\\
\textbf{CIFAR-10-P and Tiny ImageNet-P. }  
 Both these datasets contain perturbed images of the clean datasets to test the performance stability. Each example here is a short video of about 30 frames containing 10 perturbations, including noise (Gaussian and shot), blur (motion and zoom), weather conditions (snow and brightness), and affine transformations (Translate, rotation, scaling, and tilt). Frames within a video clip contain progressively increasing perturbations to challenge network stability. Unlike corruptions, stability is not measured with classification accuracy. Rather, the mean Flip Probability ($mFP$) is calculated over all perturbation categories. A flip event occurs when two consecutive frames' predictions mismatch. A lower mFP score indicates a more robust network.

 \subsection{{\mbox{Implementation and Training Details}}}\label{sec:implement}

{For an in-depth evaluation of the proposed method, we use ResNet-based \mbox{\cite{resnet1}} Wide Residual Network (WRN) \mbox{\cite{wrn1}} as the base model.} Unlike ResNet, where performance gain comes with increasing depth, WRN's architecture pivots around a central argument that shallower networks with greater width lead to higher performance gain. WRNs are denoted as \textit{WRN-d-k} where $d$ is the number of convolutional units and $k$ is the widening factor considering ResNet has a widening factor of 1, i.e., $k = 1$. We use \textit{WRN-40-2} with ReLU for the base model and replace ReLU with LP-ReLU for the proposed network. We use $Stochastic$ $Gradient$ $Descent$ or $SGD$ with $Momentum$ $0.9$. We train the network for 160 epochs with initial $learning$ $rate$ set to $0.1$. The initial learning rate is dropped by a factor of $0.2$ after 50, 100, and 140 epochs. $Batch-size$ is set to $128$ with an $L2Regularization$ factor of $.0005$. $Zero-centre$ input image normalisation is used for all experiments. 
 
 For training variants with DCT data augmentation, we follow the training protocol described in \cite{hossain1}. In DCT augmentation, each image is first transformed into the frequency domain. Later, frequency components with relatively less impact on the image are dropped based on a chosen threshold. Inverse DCT is performed afterwards with the remaining DCT coefficients for reconstructing the image in spatial domain. These reconstructed images are fed to train the deep network.

\subsection{Evaluation Metrics}

{We follow the protocol outlined in \mbox{\cite{HendrycksALP1}} for evaluating robustness against corruption and performance stability on perturbations. The protocol is briefly described below: \\
\textbf{Corruption Robustness on CIFAR-10-C and Tiny ImageNet-C. } 
 To evaluate robustness against corruption, we calculate the TOP-1 classification accuracy as in \mbox{\cite{HendrycksALP1}}. This involves validating the class prediction with the highest probability (for each test sample) against the ground truth. The ratio of correct predictions and total test samples is used as the TOP-1 accuracy.\\}
\textbf{Perturbation Stability on CIFAR-10-P and Tiny ImageNet-P. } 
Instead of TOP-1 accuracy, {as proposed in \mbox{\cite{HendrycksALP1}}} stability is measured against Flip Probability ($FP$) which provides us a quantitative insight of a network's tendency to flip prediction with increasing perturbation. To calculate $FP$, let us denote $k$ perturbation sequences (each with $v$ number of frames) with $S = \left\{\left(x_{1}^{(i)}, x_{2}^{(i)}, \ldots, x_{v}^{(i)}\right)\right\}_{i=1}^{k} $.
For a fixed $i$, i.e., perturbation type $k$, $x_{1}^{(i)}$ denotes the clean image (no perturbation) and $x_{v}^{(i)}$ denotes a frame with maximum ($k$ type) perturbation. Considering our CNN classifier as a function $F$, the Flip Probability of Network
 $F: x \rightarrow\{class_1,class_2, \ldots, class_n\}$ on perturbation sequence $S$ is:
 
\begin{equation}
F P_{p}^{F} =\frac{1}{k(l-1)} \sum_{i=1}^{k} \sum_{j=2}^{v} 1\left(F\left(x_{j}^{(i)}\right) \neq F\left(x_{j-1}^{(i)}\right)\right) 
\end{equation}
Taking the $mFP$ across all perturbations denotes the stability metric used in this work.
\\
\textbf{Clean Accuracy on CIFAR-10 and Tiny ImageNet. } 
Besides evaluating on distortions and perturbations, we evaluate the networks on the clean dataset as well. To compare the performance on clean datasets, we calculate the TOP-1 accuracy.

\begin{table}[!h]
\centering
\caption{{TOP-1 classification accuracy (\%) for different network configurations.}}

	\begin{center}
	\resizebox{1\columnwidth}{!}{%
		\begin{tabular}{c|cccc}
\hline
Network           & AF & DCT & CIFAR-10-C & Tiny Imagenet-C \\ \hline

\rowcolor{gray!25} VGG-19 &ReLU & \checkmark & 76.6 & 40.1 \\
 ResNet-101 & ReLU & \checkmark & 78.5 & 42.9 \\
\rowcolor{gray!25} WRN-40-2 & ReLU & \checkmark & 79.1 & 43.2 \\
\hline 
\hline

 VGG-19 &ReLU & \xmark & 68.5 & 35.9 \\
\rowcolor{gray!25} ResNet-101 & ReLU & \xmark & 70.2 & 37.8 \\
 WRN-40-2 & ReLU & \xmark & 72.7 & 38.6 \\
\hline 
\hline

\rowcolor{gray!25} VGG-19 &LP-ReLU\textsubscript{2} & \checkmark & 84.2 & 45.5 \\
ResNet-101 & LP-ReLU\textsubscript{2} & \checkmark & 86.5 & 47.7 \\
\rowcolor{gray!25} WRN-40-2 & LP-ReLU\textsubscript{2} & \checkmark & 89.2 & 51.9 \\
\hline
\end{tabular}
}
\end{center}

\label{table:nets}
\end{table}

\begin{table}[!h]
\centering
\caption{{TOP-1 classification accuracy (\%) for different hyperparameter configurations (Network: WRN-40-2 with \mbox{LP-ReLU\textsubscript{2} + DCT).}}}

	\begin{center}
	\resizebox{1\columnwidth}{!}{%
		\begin{tabular}{cccc}
\hline
 A, $\alpha$ & B, $\beta$  & CIFAR-10-C & Tiny Imagenet-C \\ \hline
\rowcolor{gray!25} Learnable & Learnable & 89.2 & 51.9 \\
 Learnable & Frozen & 88.1 & 50.3 \\
\rowcolor{gray!25} Frozen & Learnable & 88.4 & 50.1 \\
Frozen & Frozen & 87.9 & 49.8 \\
\hline 

\end{tabular}
}
\end{center}

\label{table:hyper}
\end{table}

\begin{table*}[!h]
\centering
\caption{TOP-1 classification accuracy (\%) on 19 individual corruptions from CIFAR-10-C and Tiny ImageNet-C with different network configurations. \textit{WRN-40-2} architecture is used in all these networks as it performs best as shown in Table \ref{table:nets}. LP-ReLU\textsubscript{2} coupled with DCT augmentation performs best. Accuracy is averaged over all five severity levels and reported in the last row. The penultimate row (Clean) reports accuracy on the respective distortion-free dataset. Best in each row is highlighted in bold. }

	\begin{center}
	\resizebox{.92\textwidth}{!}{%
		\begin{tabular}{c|ccccc|ccccc} 
\hline
&\multicolumn{5}{c|}{\textbf{CIFAR-10-C}} &\multicolumn{5}{c}{\textbf{Tiny ImageNet-C} }\\ \hline
Corruption &
  ReLU &
  LP-ReLU\textsubscript{1} &
  LP-ReLU\textsubscript{1}+DCT &
  LP-ReLU\textsubscript{2} &
  LP-ReLU\textsubscript{2}+DCT &
  ReLU &
  LP-ReLU\textsubscript{1} &
  LP-ReLU\textsubscript{1}+DCT &
  LP-ReLU\textsubscript{2} &
  LP-ReLU\textsubscript{2}+DCT  \\ \hline

\rowcolor{gray!25} Brightness      & 93.8 & 90.3  & 89.2  & \textbf{95.4}  & 94.0  & 58.3 & 58.1 & 57.3 & \textbf{59.3} & 58.6 \\
Contrast       & 74.1 & 74.2  & 75.1  & 75.3  & \textbf{80.7}  & 33.4 & 29.3 & \textbf{34.1} & 30.2 & 33.2 \\
\rowcolor{gray!25} Defocus\_blur  & 88.1 & 85.3  & 86.3  & 80.3  & \textbf{94.5}  & 44.2 & 45.9 & 52.6 & 48.6 & \textbf{56.2} \\
Elastic        & 81.3 & 79.6  & 82.5  & 79.9  & \textbf{90.5}  & 45.9 & 45.1 & 48.3 & 45.3 & \textbf{51.5} \\
\rowcolor{gray!25} Fog            & 81.7 & 84.5  & 85.2  & 85.6  & \textbf{91.2}  & 48.5 & 49.1 & 52.4 & 49.1 & \textbf{53.7} \\
Frost          & 64.4 & 74.8  & 84.6  & 75.9  & \textbf{90.4}  & 35.2 & 39.4 & 48.1 & 40.2 & \textbf{49.8} \\
\rowcolor{gray!25} Gauss\_blur    & 84.1 & 85.9  & 91.3  & 88.4  & \textbf{93.2}  & 35.6 & 38.3 & 51.6 & 40.8 & \textbf{56.7} \\
Gauss\_noise   & 51.1 & 69.5  & \textbf{88.4}  & 69.4  & 85.1  & 15.1 & 23   & 38.9 & 22.9 & \textbf{42.5} \\
\rowcolor{gray!25} Glass          & 47.8 & 53.7  & 73.8  & 53.4  & \textbf{78.6}  & 21.5 & 23.2 & 37.4 & 23.4 & \textbf{40.6} \\
Impulse\_noise & 53.6 & 78.8  & \textbf{88.3}  & 77.6  & 86.3  & 27.3 & 48.2 & 52.8 & 48.2 & \textbf{54.3} \\
\rowcolor{gray!25} Jpeg           & 75.8 & 82.6  & 86.2  & 84.5  & \textbf{94.2}  & 41.2 & 48.5 & 52.9 & 50.9 & \textbf{56.1} \\
Motion\_blur   & 80.2 & 78.3  & 82.6  & 81.2  & \textbf{90.2}  & 46.8 & 47.9 & 55.5 & 50.8 & \textbf{57.9} \\
\rowcolor{gray!25} Pixelate       & 73.6 & 72.9  & 84.1  & 73.4  & \textbf{91.7}  & 41.7 & 43.2 & 48.9 & 44.6 & \textbf{51.2} \\
Saturate       & 88.1 & 91.7  & 89.9  & \textbf{91.9}  & 91.6  & 55.4 & 56.1 & 56.4 & 56.5 & \textbf{57.7} \\
\rowcolor{gray!25} Shot\_noise    & 59.2 & 76.8  & 86.3  & 77.2  & \textbf{88.9}  & 33.5 & 45   & 52.3 & 45.2 & \textbf{55.1} \\
Snow           & 73.3 & 78.3  & 81.1  & 77.9  & \textbf{89.3}  & 45.2 & 46.7 & 52.9 & 45.3 & \textbf{55.2} \\
\rowcolor{gray!25} Spatter        & 73.5 & 84.9  & 84.6  & 85.6  & \textbf{86.9}  & 44.9 & 46.2 & 46.8 & 46.8 & \textbf{47.6} \\
Speckle        & 56.7 & 77.8  & 85.3  & 76.9  & \textbf{88.3}  & 21.5 & 33.4 & 45.2 & 33.6 & \textbf{49.9} \\
\rowcolor{gray!25} Zoom\_blur     & 80.6 & 76.2  & \textbf{92.6}  & 78.5  & 91.6 & 37.7 & 38.3 & 54.3 & 41.6 & \textbf{58.1} \\
\rowcolor{red!40}Clean          & 96.0 & 96.1 & 96.3 & 96.2 & \textbf{96.4} & 61.1 & 61.3 & 61.2 & 61.3 & \textbf{61.5} \\ \hline
  \rowcolor{blue!30} Average Accuracy &
  72.7 &
  78.7 &
  85.1 &
  79.5 &
  \textbf{89.2} &
  38.6 &
  42.4 &
  49.4 &
  43.3 &
  \textbf{51.9}\\
 \hline 

\end{tabular}
}
\end{center}

\label{table:t1_details}
\end{table*}

\begin{table}[!h]
\centering
\caption{Average TOP-1 classification accuracy (\%) comparison. LP-ReLU variants along with DCT augmentation demonstrate better robustness on both CIFAR-10-C and Tiny Imagenet-C. LP-ReLU\textsubscript{1} + DCT obtains 0.9\% and 4.8\% improvement over AugMix on CIFAR-10-C and Tiny Imagenet-C respectively. LP-ReLU\textsubscript{2} + DCT obtains 5\% and 7.3\% improvement over AugMix on  CIFAR-10-C and Tiny Imagenet-C, respectively. Best combination is highlighted in blue.}

	\begin{center}
	\resizebox{.9\columnwidth}{!}{%
		\begin{tabular}{c|cc}
 \hline
  Network           & CIFAR-10-C & Tiny Imagenet-C \\ \hline
\rowcolor{gray!25} ReLU         & $72.7\pm{0.82}$       & $38.6\pm{0.91}$            \\
Leaky-ReLU         & $67.2\pm{0.66}$       & $32.9\pm{0.72}$            \\
\rowcolor{gray!25} P-ReLU         & $68.8 \pm{0.45}$      & $33.3\pm{0.53}$            \\
Clipped-ReLU         & $73.2\pm{0.60}$       & $39.7\pm{0.66}$            \\
\rowcolor{gray!25} Tanh         & $66.9\pm{0.79}$       & $32.3\pm{0.84}$            \\
Tent         & $73.9\pm{0.29}$       & $40.5 \pm{0.33}$            \\
\rowcolor{gray!25} Cutout \cite{cutout1}       & $70.1 \pm{0.39}$       & $40.1 \pm{0.44}$           \\
Mixup \cite{mixup1}       & $72.5 \pm{0.45}$      & $40.8 \pm{0.49}$           \\
\rowcolor{gray!25} CutMix \cite{cutmix1}       & $70.8 \pm{0.36}$      & $39.7 \pm{0.40}$           \\
AutoAugment \cite{autoaugment1} & $78.8 \pm{0.34}$      & $42.5 \pm{0.37}$           \\
\rowcolor{gray!25} AugMix \cite{augmix}      & $84.2 \pm{0.28}$      & $44.6  \pm{0.30}$          \\ 

DeepAugment \cite{deepAugment}      & $78.5 \pm{0.11}$      & $42.9  \pm{0.20}$          \\ 

\rowcolor{gray!25} BNS \cite{Benz}      & $78.4 \pm{0.25}$      & $44.3  \pm{0.39}$          \\ \hline \hline

LP-ReLU\textsubscript{1}       & $78.7  \pm{0.27}$     & $42.4  \pm{0.29}$         \\
\rowcolor{gray!25} LP-ReLU\textsubscript{1} + DCT & $85.1 \pm{0.25}$      & $49.4  \pm{0.28}$          \\
LP-ReLU\textsubscript{2}      & $79.5 \pm{0.28}$      & $43.3  \pm{0.32}$          \\
\rowcolor{blue!30} LP-ReLU\textsubscript{2} + DCT & $\textbf{89.2} \pm{\textbf{0.24}}$      & $\textbf{51.9} \pm{\textbf{0.30}}$        \\ 

\hline

\end{tabular}
}
\end{center}

\label{table:t2_summary}
\end{table}

\begin{table}[!h]
\centering
\caption{{Comparison of average TOP-1 classification accuracy (\%). The combination of \mbox{LP-ReLU\textsubscript{2}} and DCT augmentation outperforms others. Best combination is highlighted in blue.}}

	\begin{center}
	\resizebox{.9\columnwidth}{!}{%
		\begin{tabular}{c|cc}
 \hline
  Network           & CIFAR-10-C & Tiny Imagenet-C \\ \hline

\rowcolor{gray!25} ReLU + DCT & $77.1 \pm{0.55}$ &  $41.2 \pm{0.61}$ \\
 Clipped-ReLU + DCT & $78.2 \pm{0.37}$ &  $42.2 \pm{0.72}$ \\
\hline \hline
\rowcolor{gray!25} LP-ReLU\textsubscript{2} + AutoAugment & $85.3 \pm{0.36}$ &  $45.2 \pm{0.28}$ \\
 LP-ReLU\textsubscript{2} + AugMix &  $86.5 \pm{0.46}$ & $47.9 \pm{0.50}$  \\
\rowcolor{blue!30} LP-ReLU\textsubscript{2} + DCT & $\textbf{89.2} \pm{\textbf{0.24}}$      & $\textbf{51.9} \pm{\textbf{0.28}}$        \\ 
\hline

\end{tabular}
}
\end{center}

\label{table:t2_aug}
\end{table}

\subsection{Comparative Performance Evaluation}
{\mbox{\textit{WRN-40-2}} architecture performs better than VGG-19 and ResNet-101 on both of the corrupted datasets (Table \mbox{\ref{table:nets}}), and therefore we use it as the backbone network for our proposed method. As Table \mbox{\ref{table:hyper}} shows, learnable hyperparameters perform better as opposed to freezing them at the initialisation values. For all our experiments, we cast the hyperparameters as learnable unless mentioned otherwise.}
\\
\textbf{CIFAR-10-C and Tiny ImageNet-C. }According to Table \ref{table:t1_details}, WRN-40-2 with ReLU as the AF achieves an average accuracy of 72.7\% over all corruptions on CIFAR-10-C. 
When LP-ReLU\textsubscript{1} replaces ReLU, overall corruption accuracy stands at 78.7\% (a jump by 6\%). A closer inspection reveals that LP-ReLU\textsubscript{1} invokes greater gains on HFc compared to the baseline (the baseline has 77.4\% and 55.1\% and LP-ReLU\textsubscript{1} has 79.5\% and 75.7\% accuracy on LFc and HFc, respectively). This complements the role low-pass filtering plays in limiting the misclassification against HFc. 

 LP-ReLU\textsubscript{2} which is specifically designed to allow greater sparsity for weaker activations, achieves 80.5\% LFc and 75.15\% HFc accuracy. 
 
 DCT augmentation complements both LP-ReLU variants by boosting overall robustness- especially against LFc. LP-ReLU\textsubscript{1} coupled with DCT augmentation achieves 84.6\% accuracy on LFc and 87\% accuracy on HFc.
LP-ReLU\textsubscript{2} coupled with DCT augmentation achieves 89.9\% accuracy on LFc and 87.1\% accuracy on HFc. 

It is worth noting that the soft phase 1 filtering in LP-ReLU\textsubscript{2} actually allows DCT augmentation to take full effect and improve performance -- especially against LFc. To be specific, the soft phase 1 filtering provides weak features (in the centre, as shown in Figure \ref{fig:main}) room to be relatively sparser. This sparsity is vital for distinguishing weak features, which otherwise congregate together. On the other hand, the hard phase 2 filtering constrains hyperactive HFc feature space and limits features from drifting away.

For a more in-depth quantitative analysis, Figure \ref{fig:grand_collage} compiles performance on each type of corruption present in CIFAR-10-C at each available corruption severity level. LP-ReLU (LP-ReLU\textsubscript{2} $+$ DCT augmentation) exhibits the best consistency across severity levels. Performance seems to drop sharply in ReLU network with increasing severity.

\begin{figure}
\begin{center}
   \includegraphics[width=1\linewidth]{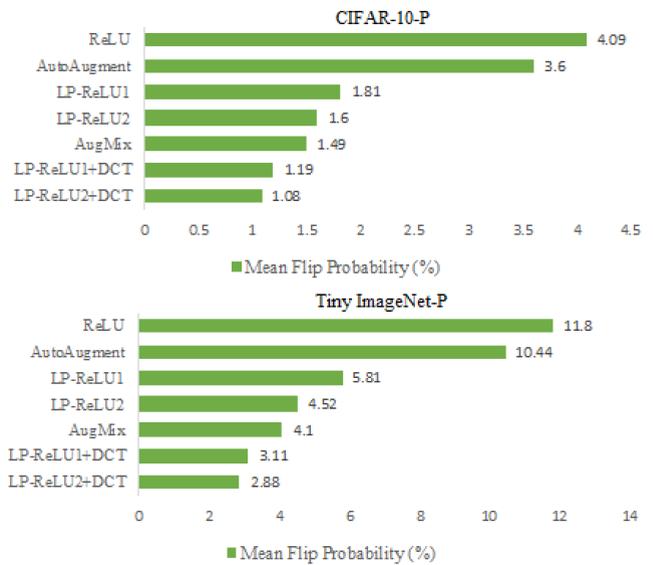}
\end{center}
   \caption{LP-ReLU\textsubscript{2} + DCT has the lowest mFP on both CIFAR-10-P (top) and Tiny Imagenet-P (bottom,) which is a testament to its performance stability across 10 perturbations present in these benchmark datasets.}
\label{fig:hor_bar}
\end{figure}

As can be seen from Table \ref{table:t2_summary}, LP-ReLU\textsubscript{2} $+$ DCT augmentation achieves SOTA Top-1 classification accuracy on both the corrupted datasets (89.2\% on CIFAR-10-C and 51.9\% on Tiny ImageNet-C). These are 5\% and 7.3\% better than AugMix \cite{augmix} on CIFAR-10-C and Tiny ImageNet-C respectively. {\mbox{LP-ReLU\textsubscript{2} + DCT} outperforms other combinations of AF and augmentation methods as shown in Table \mbox{\ref{table:t2_aug}}. We attribute the performance gain to the method's ability to address both high and low frequency corruptions}.
\\
\textbf{CIFAR-10-P and Tiny ImageNet-P. } While the corrupted datasets test the overall robustness, progressively increasing perturbations in CIFAR-10-P and Tiny ImageNet-P test the stability of network performance. Networks suffering from robustness deficiency provide erratic predictions with marginally varying perturbations. Robust networks, on the other hand, do not flip predictions unless the input corruption is substantial. As can be seen from Figure \ref{fig:hor_bar}, our proposed method, i.e., LP-ReLU\textsubscript{2} $+$ DCT augmentation, achieves SOTA $mFP$ both on CIFAR-10-P ($mFP$ 1.08\%) and Tiny ImageNet-P ($mFP$ 2.88\%). This is lower than the baseline ($mFP$ 4.09\% on CIFAR-10-P and $mFP$ 11.8\% on Tiny ImageNet-P) and the next best AugMix \cite{augmix} ($mFP$ 1.49\% on CIFAR-10-P and $mFP$ 4.1\% on Tiny ImageNet-P).\\
\textbf{CIFAR-10 and Tiny ImageNet (clean). }
 Although the proposed LP-ReLU has been designed to deal with corrupted images, both LP-ReLU variants maintain comparable results on the clean datasets with small but consistent improvements over ReLU. ReLU achieves an accuracy of 96\% on clean CIFAR-10 whereas LP-ReLU\textsubscript{1} and LP-ReLU\textsubscript{2} achieve 96.1\% and 96.2\% clean accuracy respectively. DCT augmentation slightly improves clean accuracy for both LP-ReLU variants (see Table \ref{table:t1_details}). As for Tiny ImageNet-C, similar performance gain trend is observed (see Table \ref{table:t1_details}).

We attribute our method's better robustness and stability across corruptions and perturbations to our separate approaches in handling LFc and HFc. Contemporary data augmentation-based methods \cite{augmix,cutmix1,cutout1,mixup1,deepAugment} do not account for the specific demands from corruptions residing in opposite ends of the frequency spectrum. To the best of our knowledge, only \cite{fourier1} stresses the importance of understanding the robustness issue from a Fourier perspective. Nonetheless, they resort to a pre-existing augmentation technique (AutoAugment\cite{autoaugment1}) that was originally proposed to improve clean accuracy -- not robustness. AutoAugment does improve robustness as well (as reported in \cite{fourier1}), but the improvement does not catch up with ours. AugMix \cite{augmix} further builds on AutoAugment and likewise -- do not consider the corruptions through the lens of Fourier transform. DeepAugment \cite{deepAugment} resorts to auto-encoder models for augmentation, but the synthetic data alone is not enough for significant performance gain.

\begin{figure*}
\begin{center}
   \includegraphics[width=.92\linewidth]{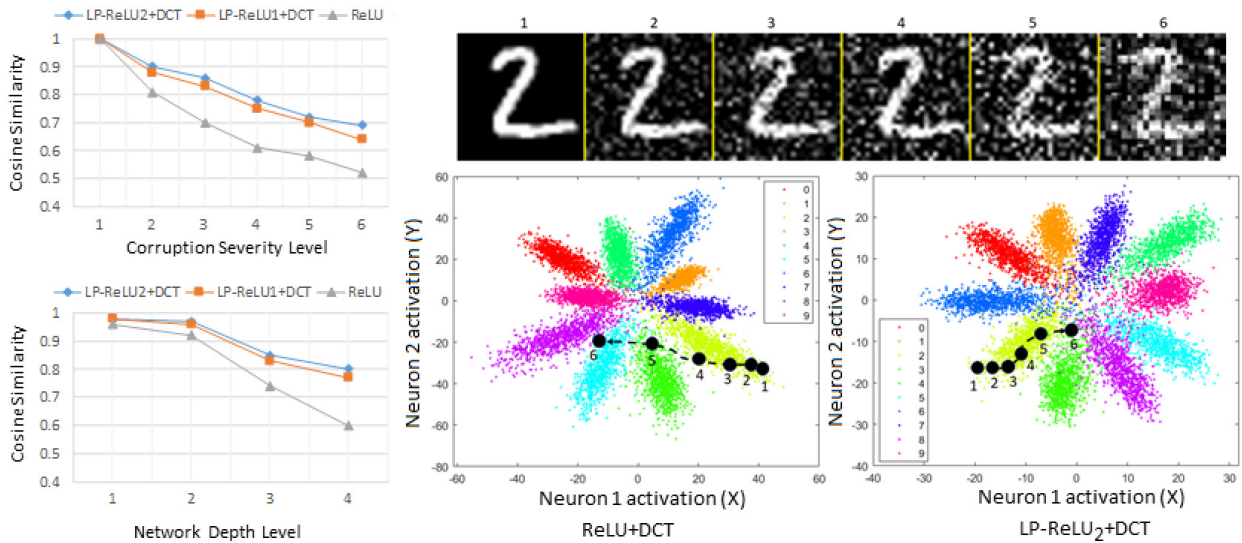}
\end{center}
   \caption{\textbf{(Top Left)} With increasing corruption severity, similarity between the clean (level $1$) and corrupted (level $[2,6]$) features fall sharply in ReLU network suggesting heavy shift in the intermediate feature space (lower similarity score refers to higher shift in feature space). Similarity score is calculated in the following severity order: $(1 \rightarrow 1), (1 \rightarrow 2), (1 \rightarrow 3), (1 \rightarrow 4), (1 \rightarrow 5), (1 \rightarrow 6$). Note that Similarity between the clean image and itself ($1 \rightarrow 1$) is 1. \textbf{(Bottom Left)}  The impact of corruption magnifies as similarity falls sharply with increasing network depth in ReLU network. Networks are divided into four depth levels ($[1,2,3,4]$) and similarity is then calculated between the corresponding depth levels, i.e., $(1 \rightarrow 1), (2 \rightarrow 2), (3 \rightarrow 3), (4 \rightarrow 4)$ on clean and corrupted datasets. \textbf{(Right)} With increasing input corruption (Gaussian noise in this case), features drift far away in ReLU network resulting in misclassification of the digit `2'. Because of a compact feature representation in LP-ReLU networks, shift in feature space is constrained and features stay close to where they belong. It is worth noticing the scale in $X$ and $Y$ axes to perceive the relative feature compactness.
   }
\label{fig:cosine_n}
\end{figure*}

\subsection{Comparing Shift In Feature Space}\label{sec:shift}

In this section, we analyse the impact of input corruption in the intermediate feature space of a CNN. To be more specific, we demonstrate how a ReLU network fares against an LP-ReLU network in terms of feature shift due to corruption.

Let us assume a CNN as a feature extractor function $\mathcal{F}(x) \rightarrow \mathbb{C}_{L}$ where $\mathbb{C}_{L}$ denotes the feature set $\mathbb{C}$ at layer $L$. The input to $\mathcal{F}$, i.e., $x$ can either be an image from the clean set $X$ ($x \in X$) or from a corrupted set $\hat{X}$. $\hat{X}$ consists of elements from $X$ that has undergone a certain corruption operation, i.e., ${Corruption}_{\Delta m}(X) \rightarrow \hat{X}$ ($x \in  \hat{X}$) where $\Delta m$ is the corruption magnitude. For a particular CNN, i.e., $\mathcal{F}$ to be robust against input corruption, $\mathcal{F}$ has to produce similar $\mathbb{C}_{L}$ for an input from $X$ and its corrupted counterpart  $\hat{X}$ (ideally exactly same) as shown in Equation \ref{eq:corruption}.

\begin{equation}\label{eq:corruption}
\begin{aligned}
\mathcal{F}(X) & \approx \mathcal{F}({Corruption}_{\Delta m}(X)), 
\quad \Delta m \rightarrow \text{small}\\
\mathcal{F}(X) & \approx \mathcal{F}(\hat{X}),
\end{aligned}
\end{equation}

{We use the Cosine Similarity (CS) \mbox{\cite{cosine_zhang}} to evaluate the effects of shift in feature space on increasing corruption severity. CS is widely used to calculate distance in high dimensional space where Euclidean or $L2$ distance does not work well \mbox{\cite{cosine_zhang}}. CS gives a measure of the angular distance between two high dimensional vectors.} We measure CS on six severity levels of corruptions as annotated in CIFAR-10-C (see Figure \ref{fig:cosine_n}). CS is measured between clean and corrupted features, one severity level at a time (average CS across layers is reported in Figure \ref{fig:cosine_n}). Level $1$ represents zero corruption, i.e., clean image and level $6$ represents maximum corruption. Compared to the baseline, both LP-ReLU\textsubscript{1} and LP-ReLU\textsubscript{2} maintain higher CS across severity levels. To put it in simple words, even in the presence of input corruption, LP-ReLU coupled with DCT augmentation produces features similar to those produced for corresponding clean images. 

In addition to different severity levels, we also measure shift at four increasing network depth levels as well. Figure \ref{fig:cosine_n} shows that our LP-ReLU based networks maintain the lowest average shift at all levels. It is also evident from Figure \ref{fig:cosine_n} that the shift is benign in the initial layers for ReLU network and becomes malignant in the deeper layers to a point where misclassification is inevitable.

As illustrated in the qualitative example in Figure \ref{fig:cosine_n}, features from a noisy digit (`2’) remain within the true class perimeter in LP-ReLU networks even with increasing magnitude of corruption. This is because of the compactness enforced by the low-pass filtering property inside LP-ReLU. However, sparsity in ReLU networks means features start drifting away even with negligible corruption resulting in misclassification.

\subsection{Training Time}\label{sec:time}

Unlike ReLU, both variants of the proposed LP-ReLU execute conditional statement during training and hence require marginally greater time per training epoch. As can be seen from Table \ref{table:time}, ReLU is the fastest with 50.3 s/epoch. LP-ReLU\textsubscript{1} turns out to be slightly faster than LP-ReLU\textsubscript{2} because of lesser parameters. However, we found both LP-ReLU networks to converge to training with fewer epochs compared to ReLU networks. E.g., ReLU requires $[180,190]$ training epochs to reach optimal performance, whereas both LP-ReLU variants perform optimally at 150 epochs. We observed that the overall training time for ReLU and LP-ReLU variants are quite similar. We argue that the sparsely represented data in ReLU requires additional training iterations to reach the global minima (with respect to the cost function). On the other hand, LP-ReLU's data representation is compact by nature and, therefore, requires fewer iterations to reach the global minima.

\begin{figure}
\begin{center}
   \includegraphics[width=1\linewidth]{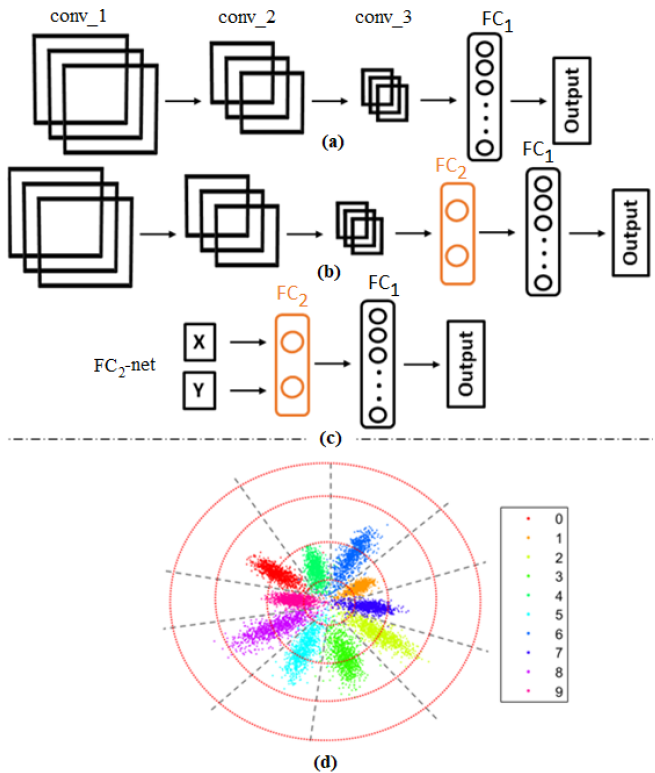}
\end{center}
   \caption{\textbf{(a)-(c)} To visualize a CNN's learned decision space on MNIST, a Fully Connected layer with 2 neurons is added as the penultimate FC layer (we call it $FC_2$). Once this network is trained, all the layers upto $FC_2$ are pruned and a tiny 3-layer network is formed (FC\textsubscript{2}-net). \textbf{(d)} Following Algorithm \ref{algo1}, $(X,Y)$ values (generated image features) are systematically fed to FC\textsubscript{2}-net and a decision space map is created from the class response.}
\label{fig:visualize}
\end{figure}

\begin{table}[!h]
\caption{Comparison of training time with different AFs (on CIFAR-10). Average time per epoch is reported in seconds. Number of epochs required to reach convergence is provided in a range. Training protocol is same as described in Section \ref{sec:implement}.}
	\label{t4}
	\begin{center}
	\resizebox{.7\columnwidth}{!}{%
		\begin{tabular}{c|ccc}
 \hline
  Network           & Time/epoch (s) & No. of epochs\\ \hline
 ReLU              & 50.3    &  $[180,190]$  \\
 Leaky-ReLU        & 55.0     &  $[185,190]$ \\
 P-ReLU              & 54.5    &  $[180,190]$  \\
 C-ReLU              & 55.9     &  $[140,145]$\\
 LP-ReLU\textsubscript{1}  & 58.1 & $[145,150]$     \\
  LP-ReLU\textsubscript{2} & 59.6  &  $[140,145]$    \\
\hline
\end{tabular}
}
\end{center}

\label{table:time}
\end{table}

 \section{Visualizing Features and Decision Space}\label{sec:visualize}

Visualising features and learned decision boundaries could deepen our understanding of CNN's robustness issues. Unfortunately, high dimensional features are hard to visualise. Dimensionality reduction algorithms based on statistical properties of features, e.g., PCA \cite{pca1} and t-SNE \cite{tsne1}, are often used in such cases. However, they do not provide an absolute feature space and the true decision space remains unknown. Multi-layer perceptrons could be used as binary classifiers for a visual explanation, but the notion of data corruption is hard to replicate in such setups. In this work, rather than using statistical properties from a high dimensional feature vector extracted from a CNN's Fully Connected ($FC$) layer, we use a two dimensional $FC$ layer to reduce the dimensions during training. This way, the network itself provides us with a lower-dimensional feature set (without using PCA or t-SNE) that is absolute in nature. In the following section, we
provide details on this feature extraction process and how we can approximate the learned decision boundary. 
 
To train a CNN on the MNIST dataset, first, we design a simple 3-convolution layer CNN that achieves $\approx 99\%$ accuracy on MNIST (Figure \ref{fig:visualize}(a)). Keeping visualization in mind, we augment another $FC$ layer ($FC_2$) right before the already existing $FC_1$ layer with only two neurons (Figure \ref{fig:visualize}(b)). {Such a low dimensional $FC$ layer is rarely used in complex tasks. However, our $FC_2$ augmented network maintains the same classification accuracy ($\approx 99\%$) on the MNIST test set which confirms the network does not over or under fit because of the additional layer. This can be attributed to MNIST's low complexity as a dataset. We argue that it is still a high dimensional dataset ($28 \times 28$ images) and it presents an opportunity to understand the decision landscape learned by CNNs trained on any image dataset.}
Plots in Figures \ref{fig:main}, \ref{fig:cosine_n}, and \ref{fig:visualize} are produced from the $FC_2$ layer features from their respective CNNs (ReLU is used in one network (ReLU-net) and LP-ReLU variants in two other networks). No traditional dimensionality reduction is used as the features themselves are only two dimensional.

\begin{algorithm}[!htbp]
\caption{: Decision Space Mapping \newline Input: Generated Features $(X_i,Y_i)$ \newline Output: Learned Decision Landscape. }
\label{algo1}
\begin{algorithmic}[1]
\STATE Init $Polar(r,\theta)$ and $i = 1$ 
\STATE where, $r \in [1:1:N], \theta \in [0:.01:2\pi]$

\FOR {$r = 1$ to $N$}
    \FOR {$\theta = 0$ to $2\pi$}
        \STATE $(X_i,Y_i) = Cartesian(Polar(r,\theta))$
        \STATE $[class_i scores_i] = classify(2FCnet,(X_i,Y_i))$
         \IF{$scores_i > 50\%$}
             \STATE $Flag_i = 1$
             \ENDIF
             \STATE $i = i + 1$
    \ENDFOR
\ENDFOR

\STATE Connect $\forall{(X_i)}\exists{(Y_i)}$ $(Flag == 1) $
\end{algorithmic}
\end{algorithm}

\subsection{Decision Space Mapping}

From the trained ReLU-net, we prune all the layers up to the $FC_2$ layer and call this tiny network FC\textsubscript{2}-net (Figure \ref{fig:visualize}(c)). FC\textsubscript{2}-net gives us the opportunity to directly mimic absolute image (digits from $0$ to $9$) features with simple two dimensional values $(X,Y)$. 

To map the decision boundary in the feature space, the following steps are followed:

\begin{enumerate}
    \item First, we plot $FC_2$ features from ReLU-net, compute the origin, and fix an initial point in the Polar co-ordinate $P(r,\theta)$ with $r = 1$ and $\theta = 0$.
    \item Next, we keep $r$ unchanged but increase $\theta$ by 0.01 up to $2\pi$ and complete a full circle traversal. We start over this process by incrementing $r$ by 1.
    \item While traversing, we move back to Cartesian co-ordinate and feed the values ($FC_2$ features) to the ReLU-net. We keep a note of the points where classification score trips over from one class to another adjacent class ($< 50\%$).
    \item Intriguingly, the class label does not alter regardless of increasing $r$ value as long as $\theta$ stays fixed. This means the decision boundaries are linear and extend to infinity.
    \item Interestingly, all inter-class trip over points from Step 2 fall in a straight line (with negligible deviation). This complements the finding in the last step, and joining these trip over points reveals the true decision boundary (see Figure \ref{fig:visualize} for a visual representation).
    
\end{enumerate}
The entire decision space mapping process is summarised in Algorithm \ref{algo1}, and visually depicted in Figure \ref{fig:visualize}(b).

While experimenting, we found that classification scores become increasingly extreme with features moving further away from the manifold. This explains why corrupted instances often induce unusually high misclassification score in ReLU-based networks as corrupted features often drift away from true class. LP-ReLU effectively reduces the open space risk by enforcing compactness and limiting sensitivity to corruptions.

\section{Conclusion}\label{sec:conclusion}

In this work, we attribute lack of robustness to unbounded Activation Functions used in modern networks. We analyse common corruptions from the frequency domain and suggest a low-pass filter based replacement for ReLU. LP-ReLU's design correlates with the Human Visual System and complies with the classic signal processing fix to HFc as well. We also suggest a data augmentation method, i.e., DCT augmentation, which appreciably complements LP-ReLU. Our proposed modifications strengthen CNN's robustness and perform consistently better across a range of corruptions. These modifications can easily be incorporated in any existing deep networks -- possibly in other deep learning domains (e.g. Natural Language Processing) as well.

{
\bibliographystyle{IEEEtran}
\bibliography{2_my_Ref}
}
 \vspace{-18 mm}
\begin{IEEEbiography}[{\includegraphics[width=1in,height=1.25in,clip]{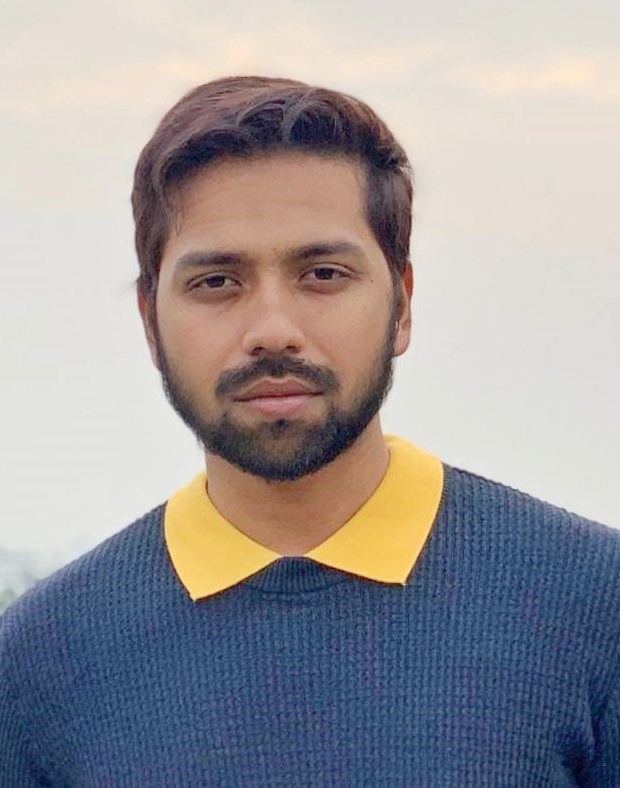}}]{Md Tahmid Hossain} is currently pursuing his Ph.D. degree at the School of Science, Engineering and Information Technology, Federation University, Australia. He received a Bachelor of Science (B.Sc.) degree in Computer Science and Engineering (CSE) from the Islamic University of Technology (IUT), Bangladesh, in 2015. He was a lecturer in CSE at IUT. His area of research interests include Computer Vision and Deep Learning: image classification, object detection, and generative adversarial networks.
\end{IEEEbiography}
\begin{IEEEbiography}[{\includegraphics[width=1in,height=1.25in,clip]{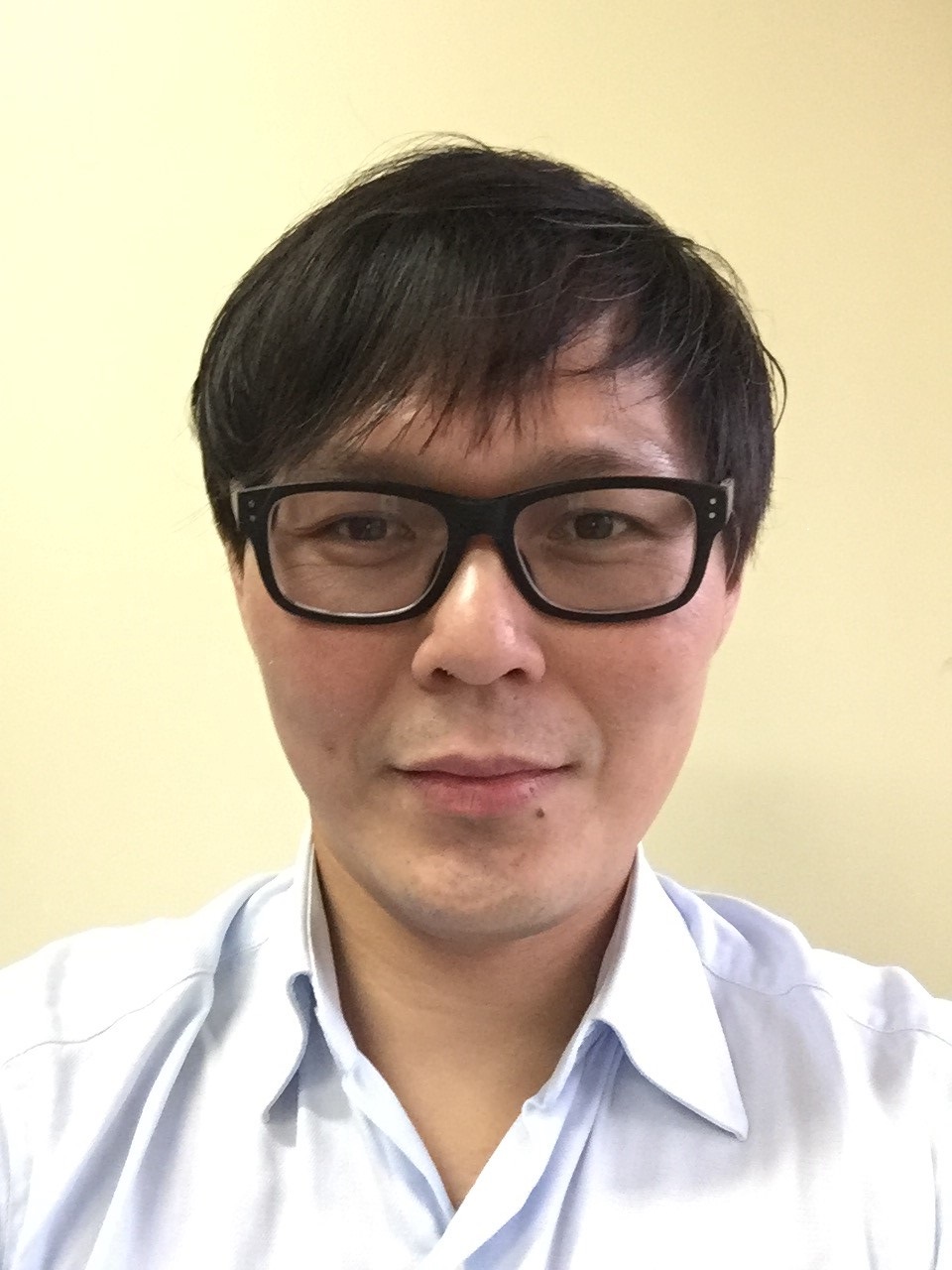}}]{Shyh Wei Teng} is an Associate Professor and Deputy Dean at School of Engineering, Information Technology and Physical Sciences, Federation University Australia. He is currently the Residential Aged Care lead of his University's partnership in the Australia's Digital Health Cooperative Research centre and a selected member of the City of Casey Smart City Advisory Committee. He previously held positions at Monash University after he obtained his PhD at that university in 2004. His research interests include Image/video processing; Machine learning; and Multimedia analytics. He has so far published over 70 refereed research papers. He received various competitive research funding, including three Federal Government grants on image retrieval and analytics—one from the Australian Research Council (ARC). He supervised 9 PhD and 1 MPhil students to completion.
\end{IEEEbiography}

\begin{IEEEbiography}[{\includegraphics[width=1in,height=1.25in,clip]{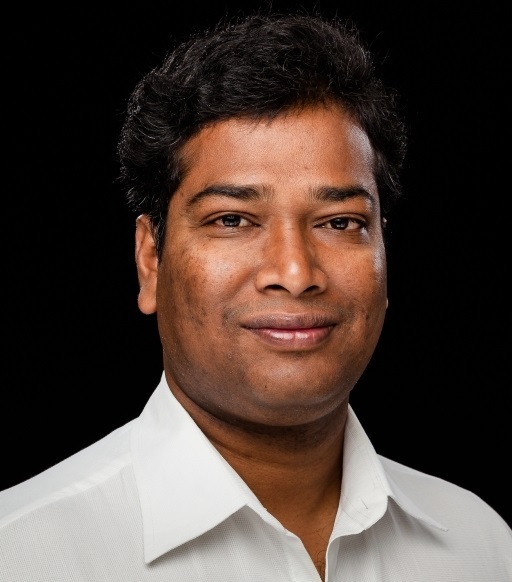}}]{ Ferdous Sohel} (M’08-SM’14) received a PhD degree from Monash University, Australia. He is currently an Associate Professor in Information Technology at Murdoch University, Australia. Prior to joining Murdoch University, he was a Research Fellow at the University of Western Australia. His research interests include computer vision, image processing, machine learning, pattern recognition, digital agriculture, and medical imaging. He is a recipient of the best PhD thesis medal from Monash University. He has received several best paper awards. He was a co-presenter of a tutorial at CVPR2015. He is an Associate Editor of IEEE Transactions on Multimedia and IEEE Signal Processing Letters. He was a Technical Program Chair of DICTA2019 and DICTA2021, Publications Chair of ICCAE2021, Organising Secretary of APCC2017, and Tutorial Chair of PSIVT2019. He is a member of the Australian Computer Society and a senior member of the IEEE.
\end{IEEEbiography}

\begin{IEEEbiography}[{\includegraphics[width=1in,height=1.25in,clip]{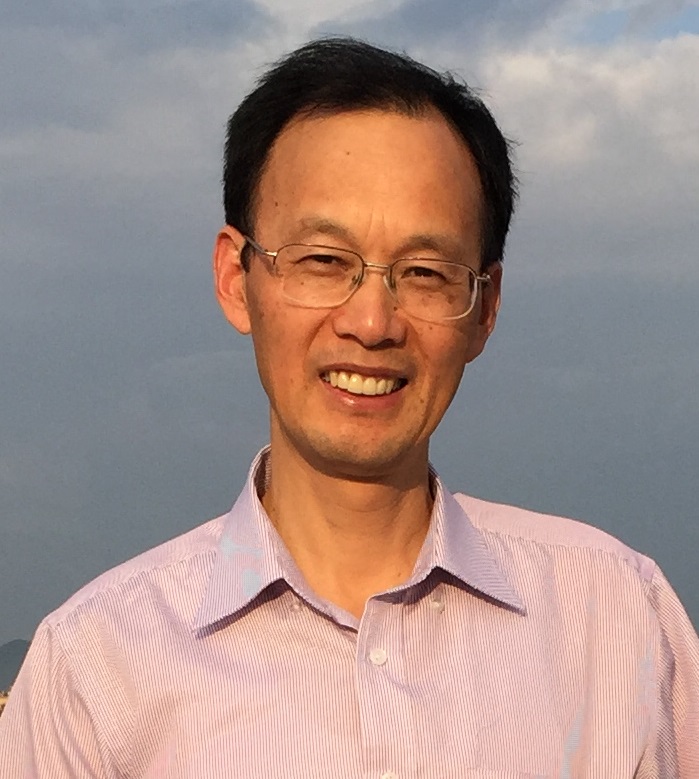}}]{ Guojun Lu} (Senior Member of IEEE) is a Professor at School of Engineering, Information Technology and Physical Sciences, Federation University Australia. He has many years’ research experience in artificial intelligence, multimedia signal processing, and retrieval, and has worked on an ARC DP project and supervising an ARC DECRA project. He has successfully supervised over 20 PhD students. He has held positions at Loughborough University, National University of Singapore, Deakin University and Monash University, after he obtained his PhD in 1990 from Loughborough University and BEng from Nanjing Institute of Technology (now South East University, China). He has published over 230 refereed journal and conference papers and wrote two books( Communication and Computing for Distributed Multimedia Systems (Artech House 1996), and Multimedia Database Management Systems (Artech House 1999)).
\end{IEEEbiography}

\EOD

\end{document}